\documentclass[letterpaper, 10 pt, conference]{ieeeconf}  

\IEEEoverridecommandlockouts                              

\usepackage{cite}
\usepackage{amsmath,amssymb,amsfonts}
\usepackage{algorithmic}
\usepackage{etoolbox}
\usepackage{graphicx}
\usepackage{subcaption}
\usepackage{textcomp}
\def\BibTeX{{\rm B\kern-.05em{\sc i\kern-.025em b}\kern-.08em
    T\kern-.1667em\lower.7ex\hbox{E}\kern-.125emX}}
\usepackage{dsfont}
\usepackage{paralist}
\usepackage{capt-of}
\usepackage{tabularx}
\usepackage{cancel}
\usepackage{bm}
\usepackage{url}
\usepackage{mathtools}
\usepackage{nccmath}
\usepackage{xcolor}
\usepackage{multirow}
\usepackage{hhline}
\usepackage{booktabs}
\usepackage{siunitx}
\usepackage{pifont}
\usepackage{comment}
\newcommand{\Agipix}{AgiPIX}
\newcommand{\Agiautonomy}{AgiAUTO}
\newcommand{\Agisim}{AgiSIM}
\newcommand{\Agiui}{AgiUI}
\newcommand{\Agihw}{AgiREAL}
\newcommand{\Agipixs}{AgiPIX }
\newcommand{\Agiautonomys}{AgiAUTO }
\newcommand{\Agisims}{AgiSIM }
\newcommand{\Agiuis}{AgiUI }
\newcommand{\Agihws}{AgiREAL }

\newcommand{\cmark}{\ding{51}}
\newcommand{\xmark}{\ding{55}}

\title{\LARGE \bf
\Agipix: Bridging Simulation and Reality in Indoor Aerial Inspection
}

\author{Sasanka Kuruppu Arachchige$^{1}$, Juan Jose Garcia$^{2}$, Changda Tian$^{3}$, Lauri Suomela$^{1}$, \\ Panos Trahanias$^{3}$ , Adriana Tapus$^{2}$ and Joni Kämäräinen$^{1}$
\thanks{*This work was supported by RAICAM, MSCA HORIZON EU}
\thanks{$^{1}$Sasanka Kuruppu Arachchige, Lauri Suomela and Joni Kämäräinen are with the Computing Sciences department at Tampere University, Finland
        {\tt\small sasa.kuruppuarachchi@gmail.com, joni.kamarainen@tuni.fi}}%
\thanks{$^{2}$Juan Jose Garcia and Adriana Tapus are with U2IS at ENSTA, Paris
        {\tt\small juan-jose.garcia@ensta.fr, adriana.tapus@ensta.fr}}%
\thanks{$^{3}$Changda Tian and Panos Trahanias is with Institute of Computer Science, Foundation for Research and Technology - Hellas (FORTH), Heraklion, Greece.
        {\tt\small dada@ics.forth.gr, trahania@ics.forth.gr}}%
}

\begin{document}
\raggedbottom
\setlength{\emergencystretch}{2em}
\sloppy
\hbadness=10000
\vbadness=10000

\graphicspath{{images/}}

\newcommand{\insertfig}{\includegraphics[width=\textwidth]{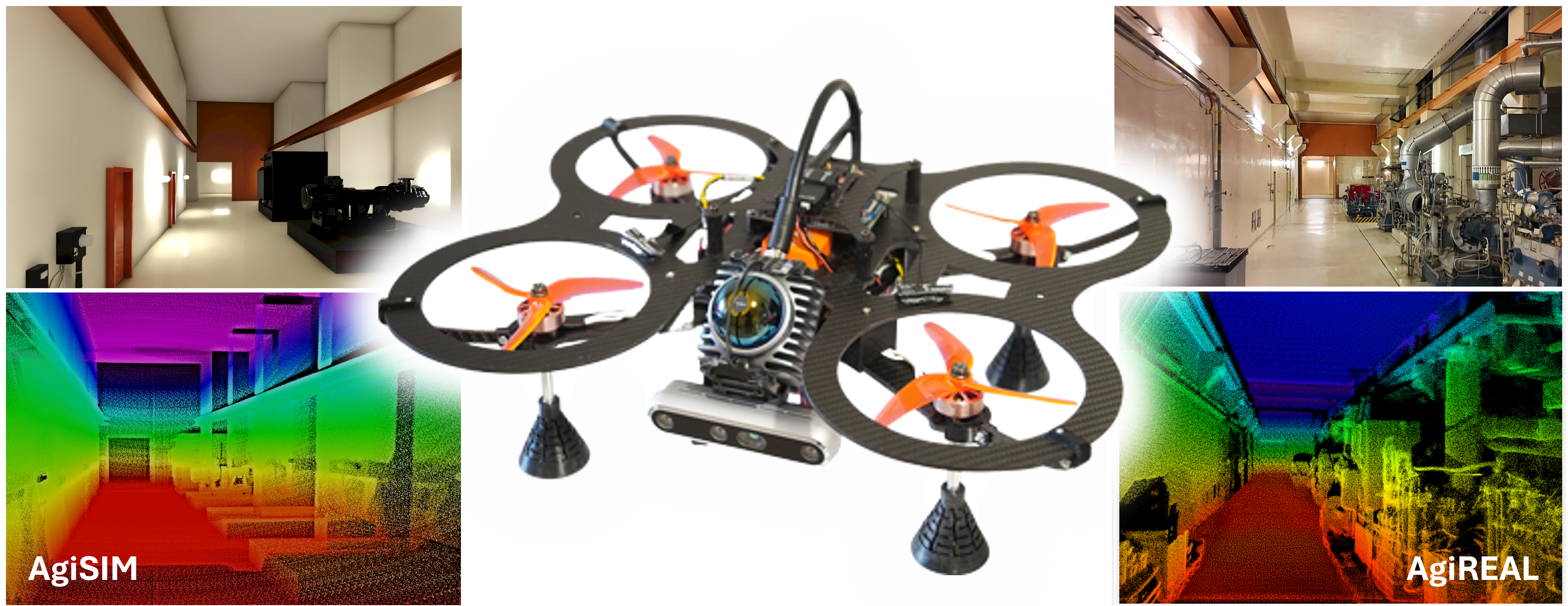}\captionof{figure}{\Agipix{} is an open platform for indoor aerial autonomy and critical asset inspection. Left: Mapping result and a PoV view of a simulated drone (\Agisim). Center: \Agipixs \Agihw::UAV hardware prototype used in field experiments. Right: Mapping result and a PoV view of a real \Agihw::UAV surveying the AKW Zwentendorf nuclear power plant, Austria.}\vspace*{-12pt}}\label{fig:coverfig}%

\makeatletter
\apptocmd{\@maketitle}{\centering\insertfig}{}{}
\makeatother

\maketitle
\setcounter{figure}{1}
\thispagestyle{empty}
\pagestyle{empty}

\begin{abstract}
Autonomous indoor flight for critical asset inspection presents fundamental challenges in perception, planning, control, and learning. Despite rapid progress, there is still a lack of a compact, active-sensing, open-source platform that is reproducible across simulation and real-world operation. To address this gap, we present \Agipix, a co-designed open hardware and software platform for indoor aerial autonomy and critical asset inspection. \Agipixs features a compact, hardware-synchronized active-sensing platform with onboard GPU-accelerated compute that is capable of agile flight; a containerized ROS~2-based modular autonomy stack; and a photorealistic digital twin of the hardware platform together with a reliable UI. These elements enable rapid iteration via zero-shot transfer of containerized autonomy components between simulation and real flights. We demonstrate trajectory tracking and exploration performance using onboard sensing in industrial indoor environments. All hardware designs, simulation assets, and containerized software are released openly together with documentation.\\
\end{abstract}
\urlstyle{same}
{\bfseries\textit{Multimedia Material}}--- \\ 
For complete documentation, code, and assets, \\ \url{https://sasakuruppuarachchi.github.io/agipix/} \relax

\section{INTRODUCTION}\label{sec:introduction}

Critical asset inspection is a key application domain for autonomous aerial robots, with growing demand across industrial facilities, energy plants, warehouses, tunnels, and other confined indoor environments. Traditional inspection is labor-intensive, costly, time-consuming, and can expose operators to safety risks in hard-to-reach or hazardous areas \cite{sanchez2024toward}. Recent advances in aerial robotics improve inspection efficiency, coverage, and safety through remote, automated data collection with onboard sensing. Commercial systems perform well outdoors, but indoor aerial inspection remains more challenging due to limited space, lack of global positioning, poor lighting, and the need for reliable close-proximity operation around obstacles and assets.

To advance the field, several research sprints have been organized, including the DARPA Subterranean Challenge \cite{tranzatto2022cerberus} for autonomous exploration in complex indoor environments and the European Robotics Hackathon (ENRICH) \cite{europeanroboticsENRICH2025} for critical asset inspection. These events enable real-world testing inside a decommissioned nuclear power plant under authentic radiological conditions. As a result, only a few research groups \cite{tranzatto2022cerberus, Foehn2022Agilicious, microswarm2022fast} have gained the expertise and resources needed to develop aerial robotic platforms, given the significant hardware and software engineering overhead.

This work aims to bridge this gap by providing an open, compact, actively sensed aerial robotics platform for indoor inspection and mapping, with a focus on reproducibility and sim-to-real transfer.

The main contributions of this paper are:
\begin{itemize}
        \item \textbf{\Agihw: Open source hardware for indoor inspection:} We present a compact, high-performance aerial platform with hardware-synchronized 3D LiDAR for precise mapping and robust navigation.
        \item \textbf{\Agiautonomy: Open source modular software:} We implement a ROS~2-based modular autonomy stack in a dockerized environment for rapid deployment.
        \item \textbf{\Agisim: Digital twin:} We provide a photorealistic Isaac Sim based digital twin~\cite{nvidiaIsaacSim2026}.
        \item \textbf{\Agiui: User interface:} We implement a low-bandwidth operator interface for mission control.
        \item \textbf{Experiments:} We report results from simulation and real-world experiments.
\end{itemize}

The paper is organized as follows: Section~\ref{sec:related} reviews related work; Section~\ref{sec:platform} presents the \Agipixs platform; Section~\ref{sec:results} reports results; Section~\ref{sec:discussion} discusses limitations and future directions; and Section~\ref{sec:conclusion} concludes.

\section{RELATED WORK} \label{sec:related}
Related work covers open aerial platforms, perception, planning, and control, simulation and digital twins, and UI.

\begin{figure*}[!ht]
    \centering
    \vspace*{-12pt}
    \includegraphics[width=1\linewidth]{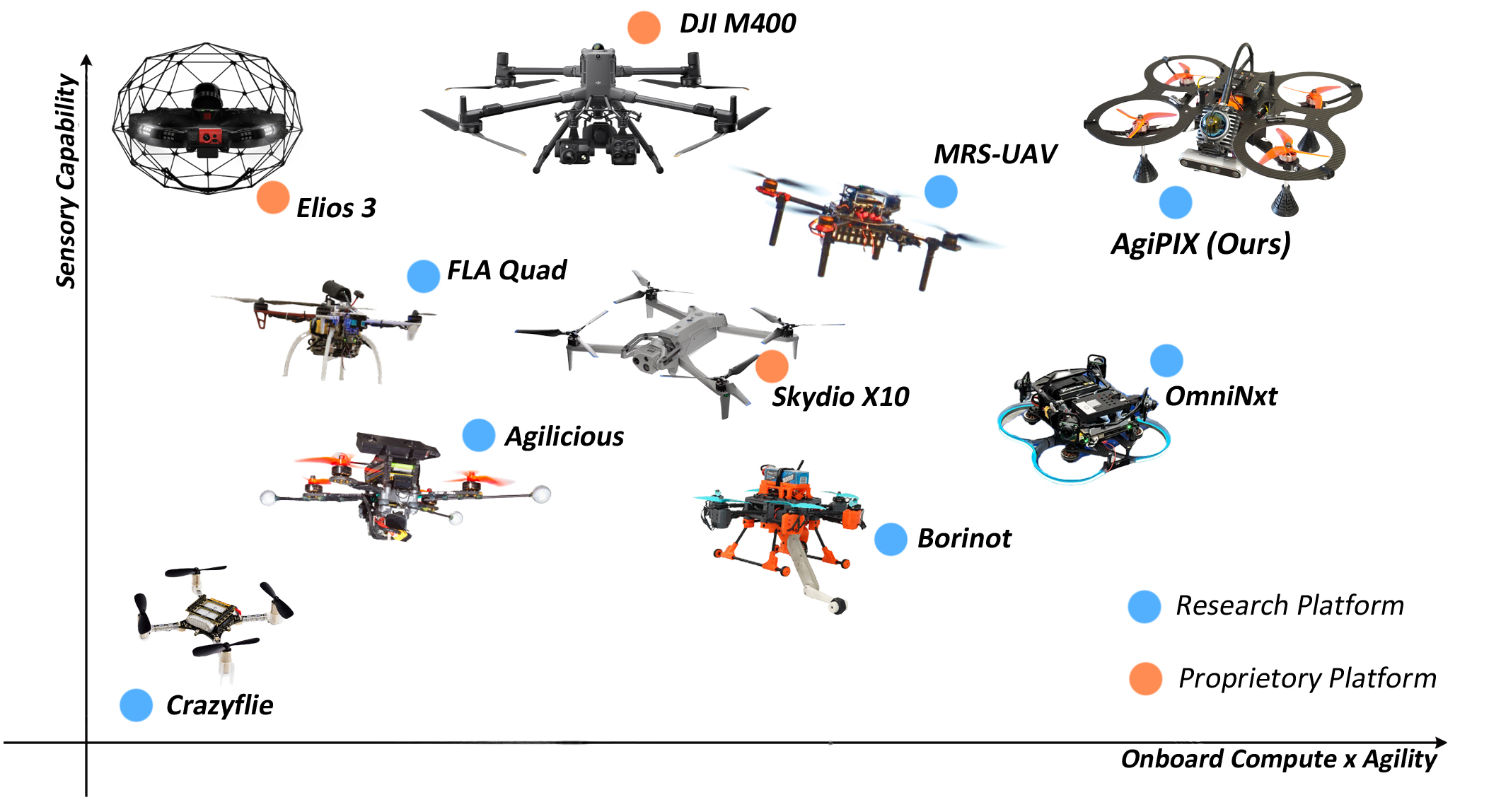}
    \vspace*{-10pt}
    
    \def\arraystretch{1.1}
	\resizebox{\textwidth}{!}{%
        \begin{tabular}{@{}l|c|c|c|c|c|r|c|c|c|r|r}
        	\toprule
        	framework & open-source & ROS~2 & simulation & \parbox{1.5cm}{\centering user interface} & \parbox{2.0cm}{\centering low-level controller} & \parbox{1.8cm}{\centering CPU Mark \\ (higher is better)} & GPU & \parbox{1.15cm}{\centering 3D LiDAR} & \parbox{1.6cm}{\centering Collision guard} & \parbox{1.3cm}{\centering thrust / weight} & \parbox{1.6cm}{\centering Diagonal span (mm)}\\
        	\midrule
        	DJI M400~\cite{djiMatrice400Specs2026} & - & \textcolor{green}{\cmark} & \textcolor{red}{\xmark} & \textcolor{green}{\cmark}  & proprietary & - & \textcolor{green}{\cmark} & \textcolor{green}{\cmark} & \textcolor{red}{\xmark} &  - & $\approx1240$\\
        	Skydio X10~\cite{skydioX102026} & - & \textcolor{red}{\xmark}  & \textcolor{red}{\xmark} & \textcolor{green}{\cmark}  & proprietary &  - & \textcolor{green}{\cmark} & \textcolor{red}{\xmark} & \textcolor{red}{\xmark} & -  & $\approx1023$\\
            Flyability~\cite{flyabilityElios32022} & - & \textcolor{red}{\xmark} & \textcolor{green}{\cmark} & \textcolor{green}{\cmark} & proprietary & - & \textcolor{green}{\cmark} & \textcolor{green}{\cmark}& \textcolor{green}{\cmark}& - & $\approx480$ \\
        	Crazyflie~\cite{GiernackiCrazy} & SW and HW & \textcolor{green}{\cmark} & \textcolor{green}{\cmark} & \textcolor{red}{\xmark} & custom & - & \textcolor{red}{\xmark} & \textcolor{red}{\xmark} & \textcolor{red}{\xmark} & $\approx2.26$  & $\approx92$\\
        	FLA-Quad~\cite{Mohta18jfr} & SW and HW & \textcolor{red}{\xmark} & \textcolor{green}{\cmark} & \textcolor{red}{\xmark} & PX4 & $3{,}383$ & \textcolor{red}{\xmark} & \textcolor{green}{\cmark} & \textcolor{red}{\xmark} & $\approx2.38$ & -   \\
        	Borinot~\cite{marti2023borinot} & SW and HW & \textcolor{green}{\cmark} & \textcolor{green}{\cmark} & \textcolor{red}{\xmark} & PX4 & $6{,}190$ & \textcolor{red}{\xmark} & \textcolor{red}{\xmark} & \textcolor{red}{\xmark} & $\approx3.50$  &  $\approx516$ \\
        	MRS UAV~\cite{Baca2021jirs} & SW and HW & \textcolor{green}{\cmark} & \textcolor{green}{\cmark} & \textcolor{green}{\cmark} & PX4 & $9{,}264$ & \textcolor{red}{\xmark} & \textcolor{green}{\cmark} & \textcolor{red}{\xmark} & $\approx2.50$ & $\approx792$  \\
        	Agilicious~\cite{Foehn2022Agilicious} & SW and HW* & \textcolor{red}{\xmark} & \textcolor{green}{\cmark}\textcolor{green}{\cmark} & \textcolor{green}{\cmark} & custom & $1{,}343$ & \textcolor{green}{\cmark} & \textcolor{red}{\xmark} & \textcolor{red}{\xmark} & $\approx5.00$ & $\approx382$  \\
            OmniNxt~\cite{liu2024omninxt} & SW and HW & \textcolor{green}{\cmark} & \textcolor{red}{\xmark} & \textcolor{green}{\cmark} & PX4 & $2{,}418$ & \textcolor{green}{\cmark} & \textcolor{red}{\xmark} & \textcolor{green}{\cmark} & $\approx4.24$  & $\approx250$  \\

                \midrule
        	\textbf{\Agipixs(Ours)} & SW and HW & \textcolor{green}{\cmark} & \textcolor{green}{\cmark}\textcolor{green}{\cmark} & \textcolor{green}{\cmark} & PX4 & $2{,}418$ & \textcolor{green}{\cmark} & \textcolor{green}{\cmark} & \textcolor{green}{\cmark}& $\approx3.50$ & $\approx495$ \\
        	\bottomrule
        \end{tabular}
        }
    \caption{
    Comparison of consumer and research platforms by onboard compute, agility, and sensing. Criteria include openness, ROS~2 support, simulation (double ticks denote photorealism), UI, CPU~performance ({\small\protect\url{https://www.cpubenchmark.net}}), and GPU availability. Agility is measured by thrust-to-weight; diagonal span is in flight configuration. (*conditionally open-source)
    }
    \label{fig:comparison}
    \vspace*{-12pt}
\end{figure*}

\subsection{Available Platforms}\label{subsec:related.platforms}
Figure~\ref{fig:comparison} summarizes the key features of \Agipixs and compares it with representative research and industrial platforms.

Commercial platforms such as DJI Matrice 400 \cite{djiMatrice400Specs2026} and Skydio X10 \cite{skydioX102026} offer mature sensing and autonomy but are large and better suited for outdoor missions. The Flyability Elios~3 \cite{flyabilityElios32022} targets confined inspection with protective structures and 3D LiDAR, yet all remain proprietary, limiting research extensibility and algorithm validation.

Open research platforms are more extensible. FLA Quad \cite{Mohta18jfr} and MRS-UAV \cite{Baca2021jirs} target GPS-denied navigation and mapping with PX4 \cite{meier2015px4}, CPU-only compute, cameras, and LiDAR, but the added payload increases size and weight, reducing thrust-to-weight ratio (TWR) and limiting close-proximity operation.

Agilicious \cite{Foehn2022Agilicious} prioritizes compactness and high TWR for aggressive flight, but its tight integration limits sensor expansion. OmniNxt \cite{liu2024omninxt} is smaller with omnidirectional perception, yet lacks 3D LiDAR, reducing mapping precision.

Across most open platforms, limited onboard GPU capability restricts real-time learning-based inspection and perception. \Agipixs combines 3D LiDAR, redundant depth and inertial sensing, and a capable GPU in a compact form factor, enabling robust indoor mapping and exploration with reproducible containerized deployment.
\subsection{Perception, State Estimation, and Mapping}\label{subsec:related.perception}
Fast, reliable state estimation underpins precise mapping. Prior work spans tightly-coupled factor-graph fusion (LIO-SAM \cite{shan2020lio}), direct point-to-map registration with efficient incremental data structures (FAST-LIO2 \cite{xu2022fast}), continuous-time trajectory estimation with higher-order motion models and observer-based stability (DLIO \cite{chen2022dlio}), and environment-adaptive LiDAR--inertial mapping with observability-aware segmentation and multi-resolution voxelization (Adaptive-LIO \cite{zhao2024adaptive}). Robustness improves with redundant perception pipelines and filtering when multiple sensors are available \cite{kuruppu2024robust}.

\subsection{Planning and Control}\label{subsec:related.planning}
\textbf{Planning:} Early quadrotor planning pipelines often relied on differential flatness and polynomial trajectory optimization (e.g., minimum-snap), which enabled smooth, dynamically feasible trajectories at high update rates \cite{mellinger2011minsnap}. Recent work expanded these foundations toward time-optimal, perception-aware local planning in clutter. In this direction, ViGO and EGO-Planner introduce ESDF-free, gradient-based local planners for agile flight in cluttered environments \cite{xuVigoPlanner2026, zhou2021egoplanner}. 

\textbf{Control:} Most real platforms still rely on a cascaded architecture with a PID-based attitude-rate inner loop for robustness and ease of tuning \cite{meier2015px4}. On top of this, advanced control approaches improve performance under constraints and disturbances, including MPC formulations \cite{lu2022onmpc} for time-optimal flight.

\textbf{Data-driven navigation policies:} Learning-based policies are increasingly used to augment or replace classical pipelines, enabling perception-aware, minimum-time flight and improved safety in cluttered scenes, with evidence of sim-to-real transfer at scale \cite{song2023perceptionaware,xu2025navrl}. Data-centric studies further quantify how synthetic vs. real data and scale impact navigation performance in unknown environments \cite{suomela2025synthetic,suomela2026data}. The modular \Agiautonomys lets \Agipixs adopt, validate, and deploy these methods in relevant scenarios.

\subsection{Simulation}\label{subsec:related.sim}
Photorealistic and physics-aligned simulators reduce development cost and risk. The aerial simulation landscape highlights that a large fraction of aerial robotics simulators are built around Gazebo, which is generally sufficient for dynamics and ROS integration but is not photorealistic \cite{furrer2016rotors}. In contrast, several simulators target higher-fidelity visuals by leveraging game engines. AirSim \cite{shah2017airsim} uses Unreal Engine, while Flightmare \cite{song2021flightmare} uses Unity to enable photorealistic rendering for perception-driven research in custom-built pipelines but lacks ROS~2 support.

The Isaac Sim ecosystem provides physics-based rendering and GPU-accelerated physics, while Pegasus enables PX4 SITL integration inside Isaac Sim \cite{jacinto2024pegasus}. \Agisims builds on this capability by implementing a digital twin of \Agipixs in Isaac Sim and sharing the same containerized autonomy stack in simulated and real flights.

\subsection{User interface}\label{subsec:related.ui}
Operator-facing UIs are typically delivered through ground control stations (GCS) that combine mission specification, real-time telemetry/health monitoring, and safety-critical handover between autonomy levels. Modern GCSs are largely map-centric, increasingly include decision-support tools, and must manage workload as supervision scales to multiple vehicles \cite{Zhang2024InterfaceWorkload}. Trends toward collaborative control, cloud-enabled thin clients, and UTM integration emphasize clear communication of automation state, intent, and constraints \cite{DiGregorio2021HumanGCS}.

In critical-asset inspection, UIs are vital for validating coverage, assessing map quality, and intervening when sensing degrades. Human-drone interaction work highlights the need to expose autonomy and safety boundaries to reduce workload and improve trust \cite{Hayat2016SurveyHDI}. These findings motivate \Agiuis as a lightweight, integrated mission-control and monitoring interface that aligns autonomy state with operator actions.

\section{\Agipixs PLATFORM}\label{sec:platform}
\subsection{System Overview} \label{sec:platform.overview}
\begin{figure*}[!ht]
        \centering
        \includegraphics[width=0.99\textwidth]{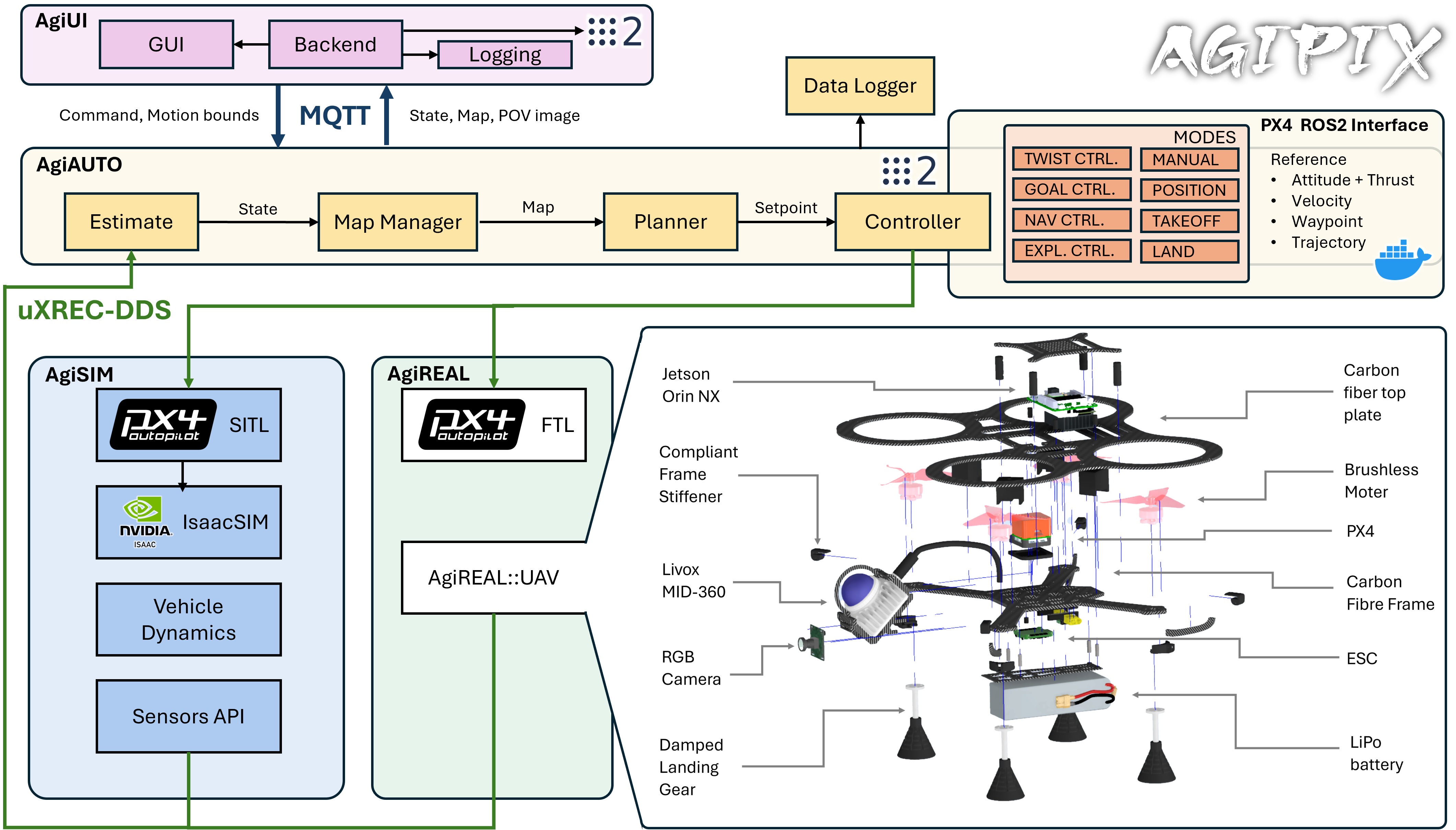}
        \caption{System overview of the \Agipixs stack. \Agiautonomys runs modular ROS~2 components on the companion computer. Flight-state and setpoint topics are bridged to PX4 via native DDS in both simulation and real flights. An external ROS~2 logger records system telemetry. \Agiuis provides operator interaction, logging, and mission commands over a MQTT \cite{holmqvist1994visually}.}
        \label{fig:sys_overview}
        \vspace*{-12pt}
\end{figure*}
\Agipixs is organized into four interacting subsystems (Fig.~\ref{fig:sys_overview}). The \Agihws hardware platform and \Agisims simulation platform share \Agiautonomys, the ROS~2 autonomy pipeline comprising estimation, a map manager, a planner, and a controller. The controller publishes setpoints and receives flight-state feedback from PX4 via native DDS. In simulation, \Agisims couples PX4 SITL with a photorealistic Isaac Sim digital twin, while in real flights PX4 runs on the \Agihw. \Agiuis provides operator interaction (GUI), backend services, and logging, and interfaces with \Agiautonomys through an MQTT bridge \cite{ika_mqtt_client2026} that carries mission commands and returns status products (state, map, and POV imagery). Finally, a logger records ROS~2 topics and telemetry for offline analysis and reproducibility.

\subsection{\Agihw \ Hardware Design} \label{sec:platform.hw}

\subsubsection{Mechanical Architecture}
The \Agipixs hardware platform is designed for safe operation near structures and within narrow passages while retaining 3D LiDAR capability. The frame constrains the diagonal motor-to-motor span to \SI{438}{\milli\meter}, with a maximum width of \SI{372}{\milli\meter} and a \SI{495}{\milli\meter} diagonal including guards. This enables passage through typical \SI{800}{\milli\meter} doors with a maneuvering margin. Propulsion is sized to yield a static thrust-to-weight ratio of 3.5:1 at full payload. The LiDAR is mounted at a 45-degree angle and protected within the carbon-fiber shell with minimal occlusion and vibration damping. Secondary sensors can be swapped between missions depending on task requirements. An overview of the components is given in Table~\ref{tab:hw_components}.
\begin{table}[!ht]
    \centering
    \footnotesize
    \def\arraystretch{1.2}
    \begin{tabularx}{\linewidth}{l|l|X}
        \toprule
        Component & \parbox{1.8cm}{Product} & Specification \\
        \midrule
        Frame & Custom open-source & \SI{4}{\milli\meter} carbon fiber \\
        Motor & T-Motor Slatts 2306 & 23$\times$\SI{6}{\milli\meter} stator, \SI{2400}{\kilo\volt}, \SI{758}{\watt}  \\
        Propeller & \parbox{1.8cm}{Azure Power SFP5148} & \SI{5.1} inch length and \SI{4.8} inch pitch \\
        Battery & Tattoo G-Tech 4500 & 6$\times$ \SI{3.7}{\volt}, \SI{4500}{\milli\ampere\hour} \\
        Flight Controller & Pixhawk Orange & Built in redundancy \\
        Motor Controller & F55A Pro II 3-6S & DShot protocol, 4$\times$ \SI{60}{\ampere} \\
        Compute Unit & \parbox{2.0cm}{nVidia Jetson Orin Nx (Super)} & 8$\times$ A78 \SI{2.0}{\giga\hertz}, \SI{16}{\giga\byte}, 157 TOPS \\  
        \midrule
        LiDAR & Livox MID 360 & \SI{50}{\meter} at 360°$\times$59° FOV \\
        IMU & Pixhawk Orange & Isolated and triple redundant\\
        Optical flow & HereFlow PX4 & Redundant velocity measurement \\
        Depth (Opt.) & RealSense D455 & \SI{6}{\meter} m at 87°$\times$58° FOV \\
        RGB (Opt.) & Arducam OG02B10 & Global Shutter \\
        Radiation (Opt.) & DFrobot Gravity & Ionizing Radiation Detector \\
        \bottomrule
    \end{tabularx}
    \caption{Overview of the components of the flight hardware design.}
    \label{tab:hw_components}
    \vspace*{-15pt}
\end{table}
\subsubsection{Sensors and Hardware Synchronization}
\begin{figure}[!ht]
        \centering
        \includegraphics[width=0.99\linewidth]{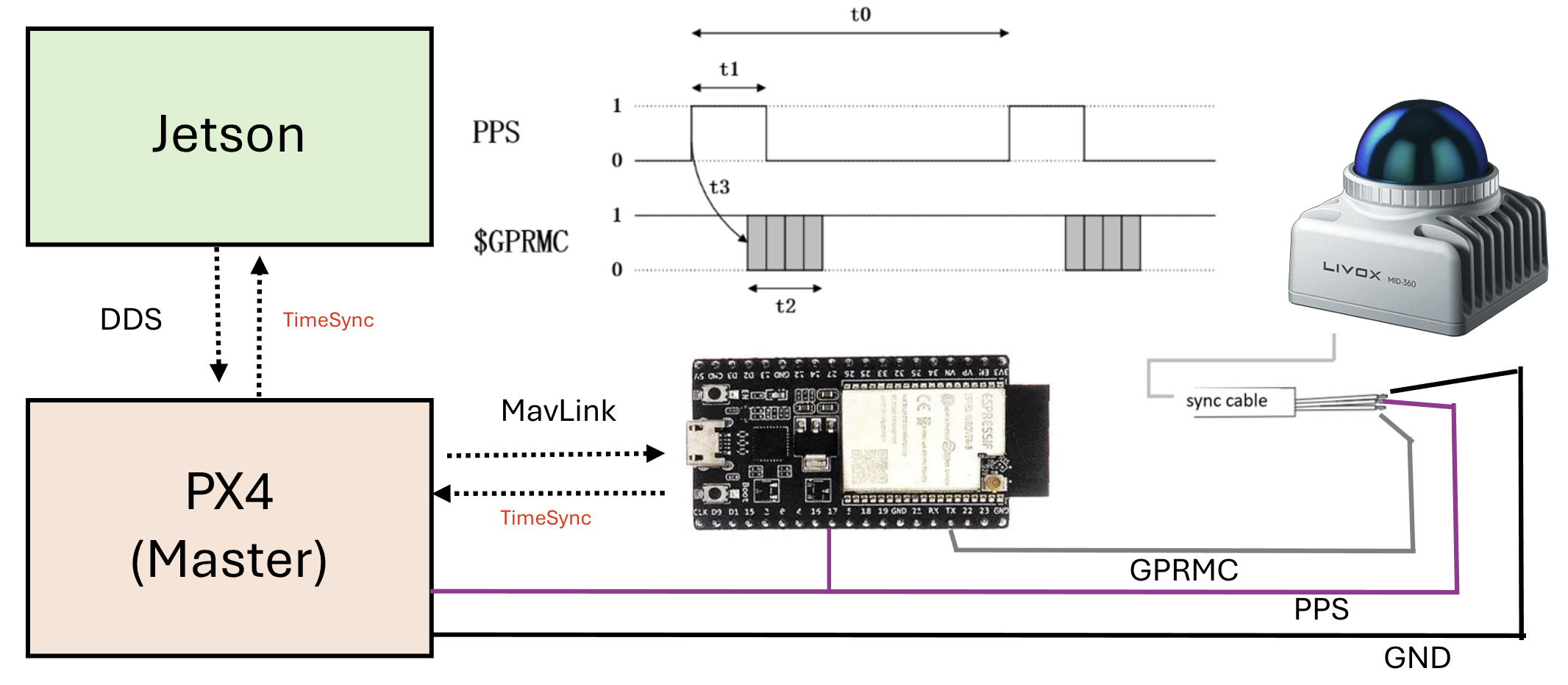}
        \caption{Hardware time synchronization pipeline. The Pixhawk (IMU) provides the master clock; the LiDAR is synchronized via PPS, with an ESP32 generating the required timing signal.}
        \label{fig:hw_sync}
        \vspace*{-12pt}
\end{figure}

The primary sensors of the platform are the LiDAR and IMU, which are used for LiDAR--inertial odometry (LIO) state estimation. Accurate synchronization between these sensors is critical for reliable estimation. We achieve this using the pulse-per-second (PPS) synchronization capability of the Livox Mid-360 LiDAR. The Pixhawk, which houses the IMU, acts as the master clock, and both the companion computer and LiDAR are synchronized to it. An ESP32 connected to the Pixhawk via MAVLink generates the timing signal required by the LiDAR. An overview of the method is shown in Fig.~\ref{fig:hw_sync}. USB~3.0, CAN, UART, RS-232, and SPI interfaces are available for optional secondary sensors.

\subsection{\Agiautonomy: Modular Software Framework} \label{sec:platform.sw}
\Agiautonomys is organized as a set of ROS 2 packages and runtime services that communicate over ROS 2. To enable reproducible deployment, \Agiautonomys lives in a container image and is launched through a unified orchestration. This isolates dependencies, makes experiments portable across compute targets, and allows switching between sim and hardware by changing only the configuration

\noindent\textbf{Middleware and interfaces:} We use Micro XRCE-DDS \cite{PX4_ROS2_bridge} to interface directly with PX4 uORB topics. Safety-critical primitives remain in PX4, while higher-level autonomy runs onboard the companion computer. A thin interface layer \cite{PX4_ROS2_Control_Interface} translates between ROS~2 messages and validated PX4 setpoint streams.

\subsubsection{Perception: State Estimation and Mapping}
\Agiautonomys provides a LiDAR--inertial state-estimation and mapping pipeline for GPS-denied indoor environments. The default configuration uses a modified Adaptive-LIO \cite{zhao2024adaptive} with an EKF that fuses the IMU to improve robustness and update rate.

For navigation in cluttered industrial sites, the perception stack flags dynamic obstacles using LV-DOT \cite{LV-DOT}. The local environment is maintained as a 3D occupancy structure and the global map as a voxel grid. These feed the planner and \Agiui, while higher-resolution maps are recorded for offline inspection.

\subsubsection{Planning and Control} \label{sec:platform.sw.plan_ctrl}
\Agiautonomys exposes multiple flight modes that separate operator intent, autonomy level, and safety constraints, enabling staged bring-up while reusing PX4 safeguards. Beyond PX4's standard modes, the stack provides four autonomy modes: \emph{Twist control} (direct velocity commands from \Agiui), \emph{Goal control} (\Agiuis controlled goal pose with obstacle avoidance), \emph{Navigation control} (global waypoint/trajectory following), and \emph{Exploration control} (closest-frontier autonomy).

Trajectory generation uses a local ViGO planner \cite{xuVigoPlanner2026} that produces dynamically feasible polynomial trajectories in the current occupancy map. The resulting trajectory is tracked by an on-manifold MPC \cite{lu2022onmpc}, which outputs smooth attitude/thrust references streamed to PX4 as trajectory setpoints at a fixed rate.

\textbf{Data-driven navigation policies:} The modular design and separated control modes of \Agiautonomys enable learning-based navigation. Following Section~\ref{subsec:related.planning}, we evaluate a learned policy by deploying the Fast Appearance-Invariant Navigation Transformer (FAINT) \cite{suomela2025synthetic} onboard \Agihw. The policy, trained in simulation, transfers zero-shot to the platform, illustrating the adaptability enabled by \Agiautonomy.

\subsubsection{Data Logging}
Reliable data logging is mission-critical for inspection tasks. ROS~2 bags can grow quickly, become corrupted, and consume system memory. We provide \textit{agi\_logger} \cite{Kuruppuarachchi2026agi_logger}, an open-source package for reliable logging and TCP file transfer that uses \texttt{.mcap} storage with autostart, memory/time limits, and lossless segmented recording. \texttt{.mcap} log files enable offline inspection using tools such as Foxglove \cite{foxglove2026}.

\subsection{\Agisim: Digital Twin} \label{sec:platform.sim}

\subsubsection{Simulation Environment and Workflow}
\Agisims provides a photorealistic digital twin if the \Agihw. As summarized in Fig.~\ref{fig:sys_overview}, Isaac Sim renders the environment and simulates the onboard sensors with time-stamped outputs. PX4 software-in-the-loop (SITL) is handled through the Pegasus interface \cite{jacinto2024pegasus} communicates via DDS, so that the same ROS~2 topics are available to \Agiautonomys as in real flight.

\noindent\textbf{Workflow.} A typical experiment proceeds as follows: (i) launch Isaac Sim with the scene, robot, sensor configuration and PX4 SITL, (ii) start the ROS~2 bridge, (iii) launch the \Agiautonomys containers, and (iv) visualize and supervise the run via \Agiui/Foxglove. The same mission definitions are used across sim and real flights, which supports rapid iteration while keeping deployment consistent.

\subsubsection{Reproducibility and Configuration Parity}
To reduce sim-to-real drift, \Agisims enforces configuration parity at three levels. 
\emph{Sensor parity} is maintained by matching calibration and noice parameters frames and topic conventions with the hardware setup. 
\emph{Timing parity} is achieved by propagating simulation time consistently to ROS~2. 
\emph{Software parity} is ensured by running the exact same container images and launch files in both environments.

\begin{figure*}[t]
  \centering
  \includegraphics[width=\textwidth]{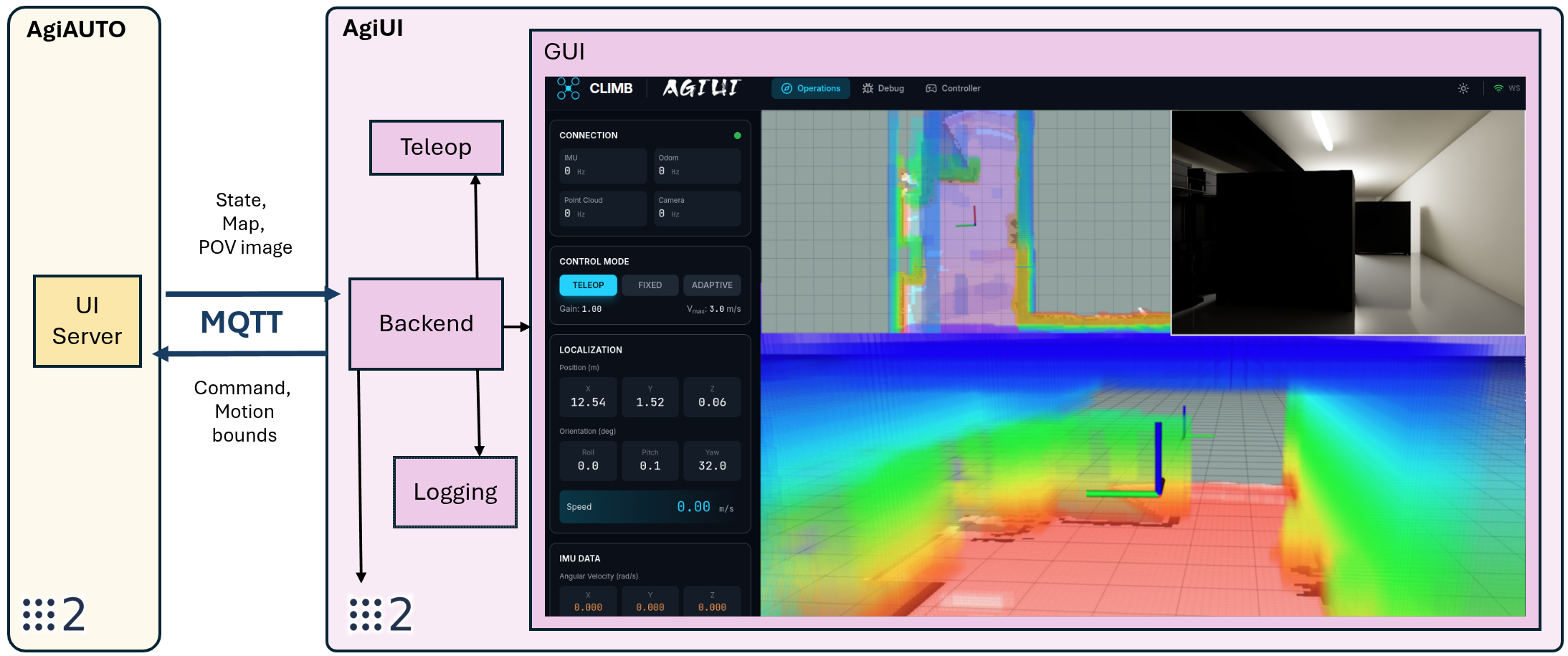}
  \caption{This is the complete overview of the AgiUI where the user can see different perspectives of the robots it is controlling, including a top view, a side view, and a third-person view.}
  \label{fig:ui_complete_agiu}
\end{figure*}

\subsection{AgiUI: User Interface}\label{sec:platform.ui}

\Agiuis is the operator-facing interface that closes the loop between mission intent and the onboard autonomy stack (Fig.~\ref{fig:sys_overview}). We implement \Agiuis as a lightweight web application that bridges commands and status through a low-bandwidth MQTT channel \cite{Hu2021CloudStation}. The UI server receives state, map updates, POV imagery, and payload readings over the MQTT channel. User commands and motion bounds are sent back to be received by the robot-side UI backend, which translates them back to ROS~2 topics.

The GUI is organized around (i) a fleet overview and (ii) synchronized spatial products. The fleet panel lists all active robots under supervision using a common abstraction of connection state, autonomy mode, and basic health indicators. A 3D mapping view provides a shared, joint reconstruction from the different robots, enabling operators to inspect explored regions, verify coverage, and contextualize robot poses and trajectories. For radiological inspection, \Agiuis additionally renders a radiation map aligned with the shared environment representation, supporting rapid identification of hotspots and informing viewpoint replanning. Per-robot control tabs expose mission specification (geofences, waypoints, and inspection tasks) and teleoperation primitives, with explicit mode switching (manual/assisted/autonomous) to support safe handover procedures.

Beyond robot control, \Agiuis includes a Human-Robot Interaction (HRI) monitoring panel to support user-centric supervision. Building on our prior work on cognitive-load dynamics in teleoperation \cite{GarciaCardenas2024CognitiveLoadTeleop}, this panel can ingest physiological streams (e.g., heart-rate variability, electrodermal activity/GSR, and eye-based measures) and display online indicators of cognitive load and trust alongside mission context. These signals are logged with the rest of the system telemetry, enabling post-hoc analysis and future closed-loop policies that adapt autonomy level and information presentation to the operator's state.

\section{RESULTS}\label{sec:results}
This section summarizes inspection- and mapping-focused experiments conducted in both simulation (\Agisim) and real-world flights and outlines the evaluation protocol used during ENRICH~2025.

\begin{table}[!ht]
    \centering
    \footnotesize
    \def\arraystretch{1.2}
    \begin{tabularx}{\linewidth}{l|r|r}
        \toprule
        \parbox{3.3cm}{\textbf{Trajectory}} 
        & \parbox{2cm}{\centering\textbf{AgiSIM ATE (m)}} 
        & \parbox{2cm}{\centering\textbf{AgiREAL ATE (m)}}\\
        \midrule
        Lemniscate  & 0.0637 & 0.1404 \\
        Up Down Spiral  & 0.0368 & 0.1049 \\
        \bottomrule
    \end{tabularx}
    \caption{Trajectory-tracking performance comparison between AgiSIM and AgiREAL across different trajectories.}
    \label{tab:rmse_results}
    \vspace*{-12pt}
\end{table}

\begin{figure}[t]
        \centering
        \includegraphics[width=0.98\linewidth]{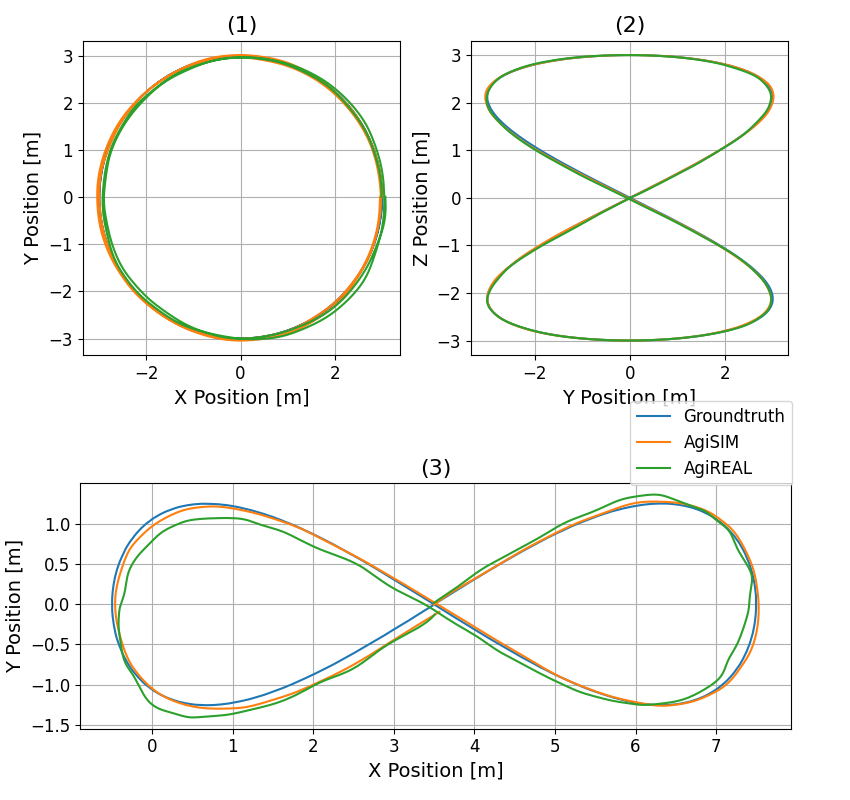}
        \caption{Trajectory-tracking comparison of Ground Truth, \Agisim, and \Agihws in two trajectories at 6 m/s maximum velocity; Up down spiral and lemniscate: (1) spiral in the $x$--$y$ plane, (2) spiral in the $y$--$z$ plane, and (3) lemniscate in the $x$--$y$ plane.}
        \label{fig:result_tracking}
        \vspace*{-13pt}
\end{figure}
\subsection{Trajectory tracking performance}\label{subsec:results_trajectory}
We evaluate trajectory-tracking accuracy in both \Agisim{} and \Agihw{} using two representative paths: a lemniscate and an up--down spiral (Fig.~\ref{fig:result_tracking}). As described in Sec.~\ref{sec:platform.sw.plan_ctrl}, the predefined polynomial trajectories are tracked by an on-manifold MPC \cite{lu2022onmpc} that produces attitude setpoints for the low-level controller. Real-world poses are measured by a VICON system. Each run is executed at a maximum linear speed of \SI{6}{\meter\per\second} and a maximum linear acceleration of \SI{5}{\meter\per\second}, and tracking performance is reported as ATE RMSE with respect to the ground-truth trajectory (Table~\ref{tab:rmse_results}).

Across both trajectories, \Agipix{} maintains sub-decimeter tracking error in simulation and low-decimeter error on hardware. In \Agisim{}, RMSE is \SI{0.0637}{\meter} (lemniscate) and \SI{0.0368}{\meter} (up--down spiral); on \Agihw{}, RMSE is \SI{0.1404}{\meter} and \SI{0.1049}{\meter}, respectively. Despite the expected sim-to-real gap from unmodeled disturbances and sensing/actuation effects, tracking remains stable under fast direction changes and altitude variations.

\begin{figure}[ht!]
        \centering
        \includegraphics[width=0.99\linewidth]{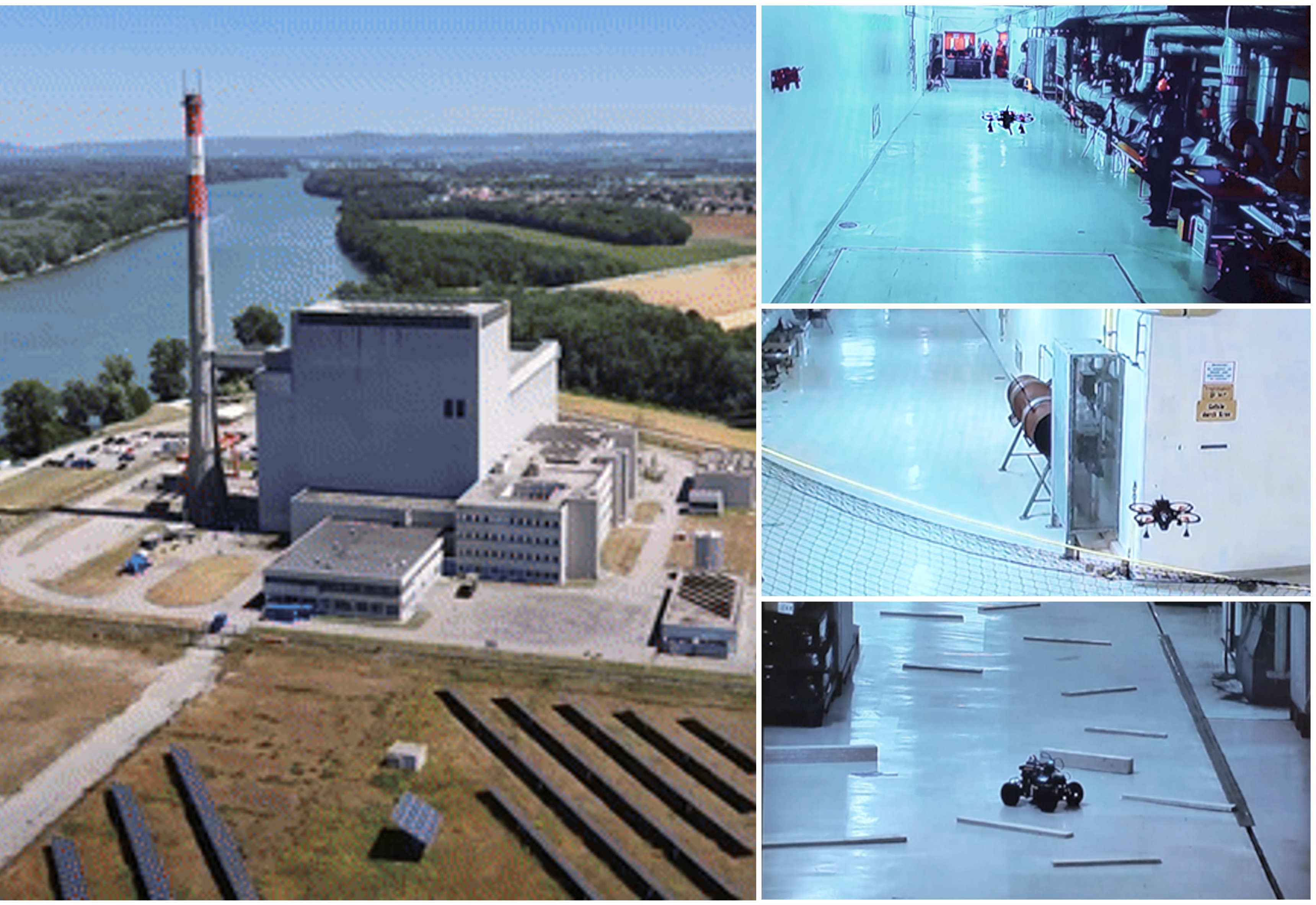}
        \caption{Left: AKW Zwentendorf, Right: \Agipix during the various stages of the trial.}
        \label{fig:enrich_team}
\end{figure}
\begin{figure}[t]
        \centering
        \includegraphics[width=0.98\linewidth]{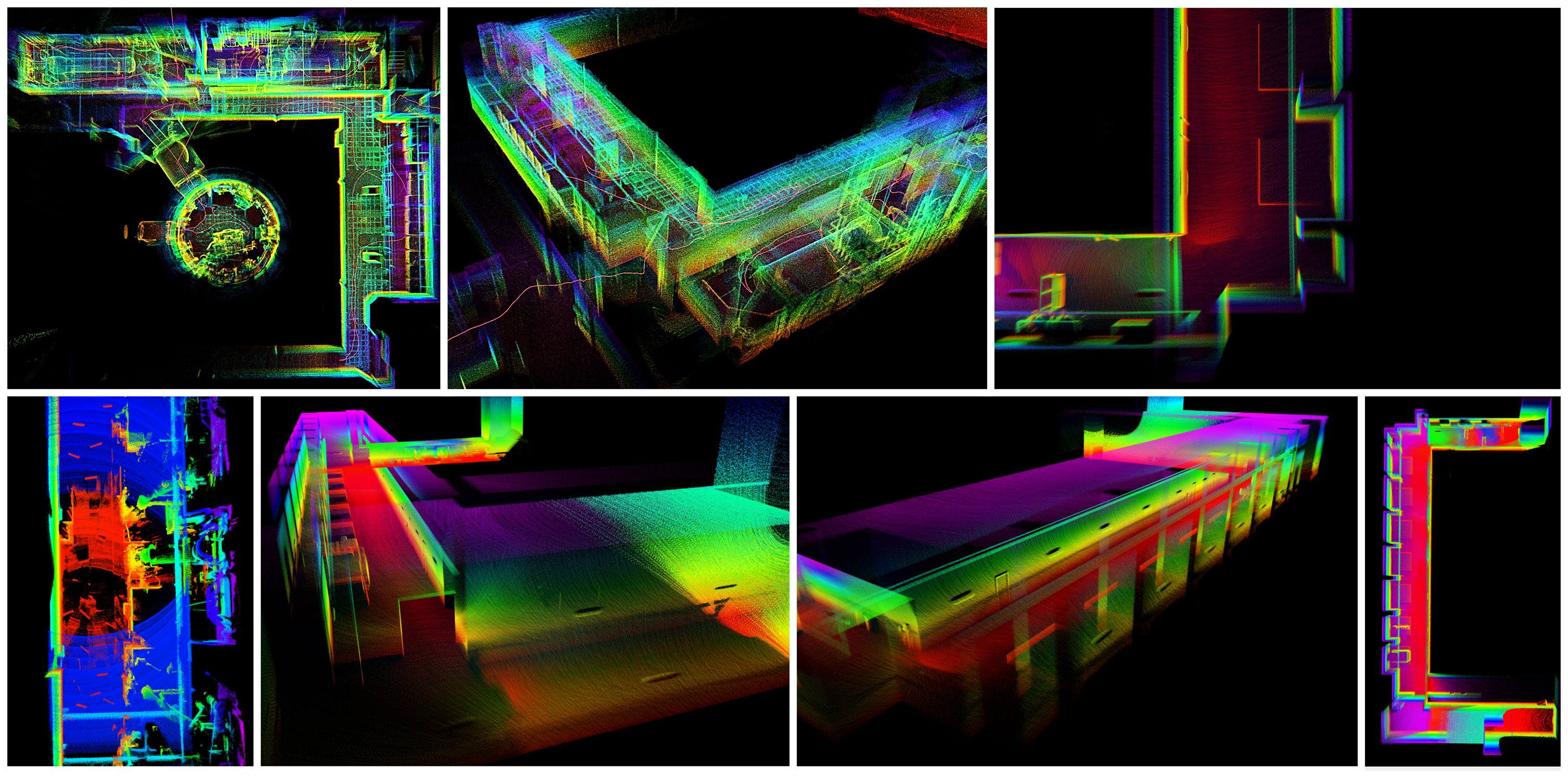}
        \caption{Representative ENRICH~2025 mapping output.}
        \label{fig:enrich_mapping}
\end{figure}
\begin{figure}[t]
        \centering
        \includegraphics[width=0.98\linewidth]{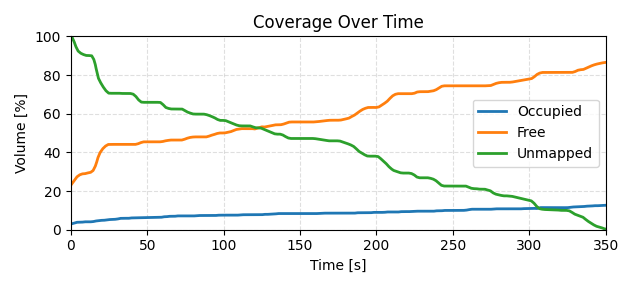}
        \caption{Percentage of free, occupied, and unmapped volume over the total exploration time for the exploration mode using the closest-frontier method for the given map}
        \label{fig:result_explore}
        \vspace*{-6pt}
\end{figure}

\subsection{Mapping Results}\label{subsec:results.mapping}
We report mapping fidelity (IoU) and exploration coverage over time (Fig.~\ref{fig:result_explore}) in a representative indoor environment. The experiment uses a modified Adaptive-LIO \cite{zhao2024adaptive} with Closest Frontier planning \cite{xu2021CFExplore}. We achieved an IoU of 0.96 between the occupancy grid and the ground-truth map over the explored region.

We validated \Agipixs at ENRICH~2025~\cite{europeanroboticsENRICH2025} on a large-scale indoor mapping task (Fig.~\ref{fig:enrich_team}). The \Agihw::UAV runs were complemented by a \Agihw::UGV for extended floor-level coverage. The mission reached 72\% of the planned flight distance before a communication failure; Fig.~\ref{fig:enrich_mapping} shows the resulting map.

\begin{figure}[!ht]
    \centering
    \includegraphics[
    trim={3cm 5cm 3cm 3cm}, clip,
    width=\linewidth]{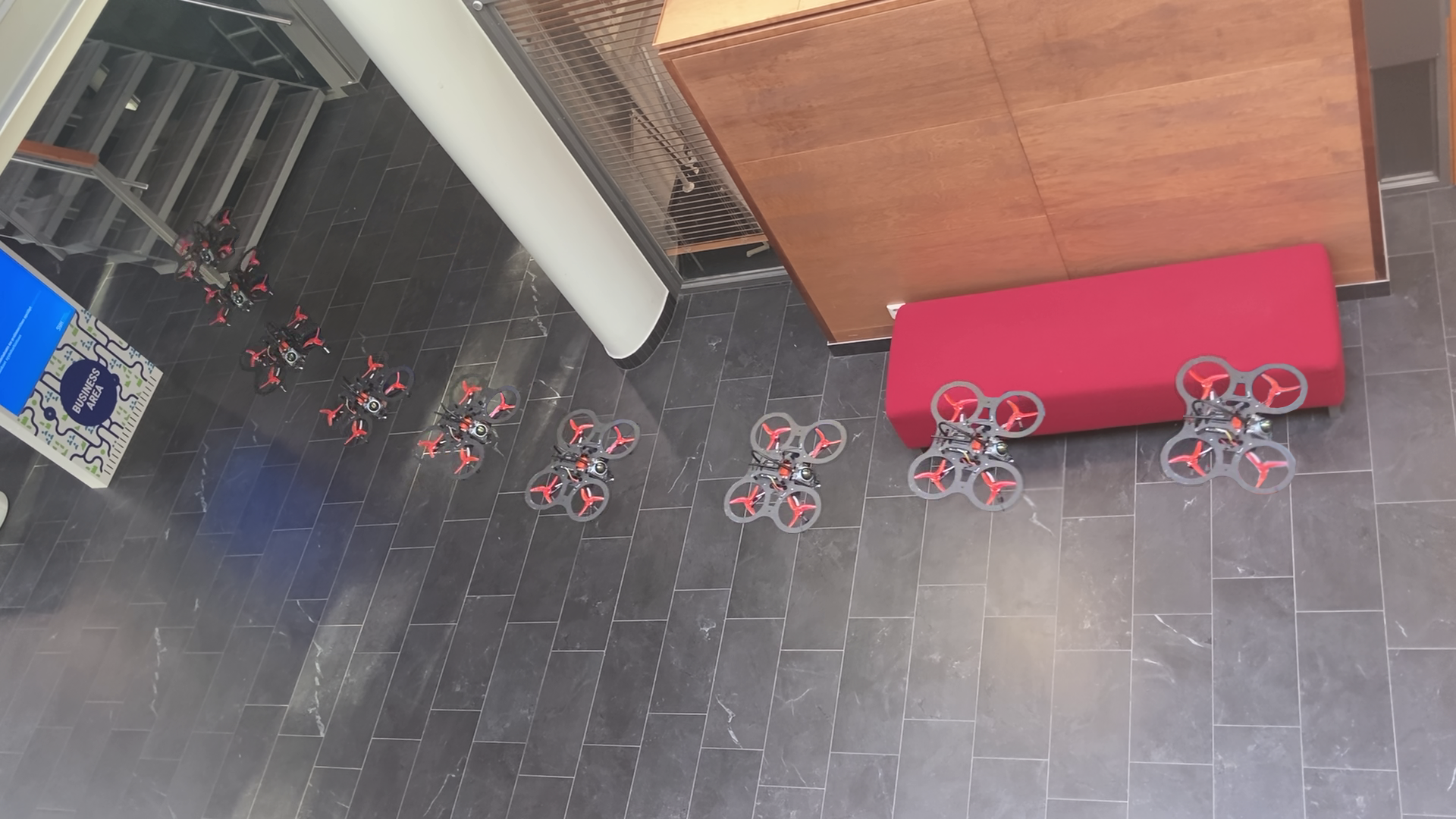}
    \centering
    \includegraphics[
    trim={0cm 0.3cm 0cm -0.4cm}, clip,
    width=1.0\linewidth]{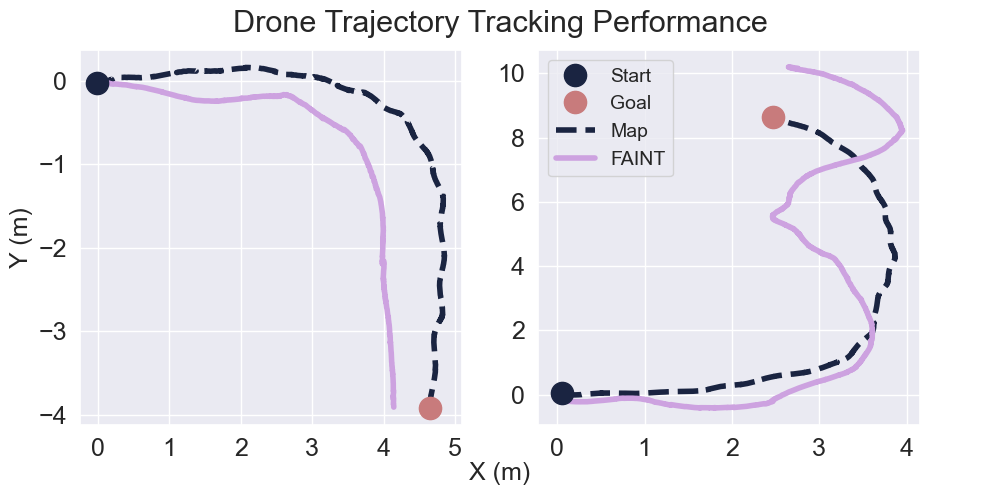}
    
    \caption{ Drone trajectory relative to the mapping trajectory when controlled by FAINT \cite{suomela2025synthetic}, a learning-based visual navigation method. The drone successfully reached the goal on the left trajectory, but missed it on the right.}
    \label{fig:drone_trajectories}
    \vspace*{-12pt}
\end{figure}

\subsection{Learning-Based Navigation Results}\label{subsec:results.learning}
We evaluated a learned visual navigation policy on the \Agipixs drone without retraining. The FAINT model \cite{suomela2025synthetic} was deployed onboard to control forward velocity and yaw rate at fixed altitude. In real-wprld experiments (Fig.~\ref{fig:drone_trajectories}), the policy transferred across embodiment, achieving \SI{1.07}{\meter} RMSE on a successful run and \SI{1.42}{\meter} RMSE on a more challenging route, with \SI{120}{\milli\second} inference latency on the Jetson Orin NX.

These results show \Agipix's ability to run state-of-the-art learning-based methods that require fast visual feature computation, and to serve as a validation platform for learning-based navigation.

\subsection{System Utilisation on the Onboard Compute Unit}

Table \ref{tab:utilization_table} reports the average CPU and GPU utilisation of \Agiautonomy\ on the Jetson Orin NX Super 16 GB. The system uses 41\% of the CPU and 37\% of the GPU, leaving 59\% and 63\% free resources, respectively.

Overall, the results indicate substantial computational headroom, supporting real-time operation and future scalability.

\begin{table}[!ht]
    \centering
    \footnotesize
    \def\arraystretch{1.2}
    \begin{tabularx}{\linewidth}{l|r|r}
        \toprule
        \parbox{3.3cm}{\textbf{Software Component}} & \parbox{2cm}{\centering\textbf{CPU Utilisation (\%)}} & \parbox{2cm}{\centering\textbf{GPU Utilisation (\%)}}\\
        \midrule
        AdaptiveLIO & 13 & 5 \\
        Livox Driver & 9 & 2 \\
        Map Manager & 4 & 20 \\
        Ego planner & 1 & 10 \\
        Autonomous flight & 10 & 0 \\
        Controller & 1 & 0 \\
        PX4 ROS~2 Interface & 1 & 0 \\
        uXREC-DDS & 1 & 0 \\
        \midrule
        \textbf{Free Resources} & \textbf{59} & \textbf{63} \\
        \bottomrule
    \end{tabularx}
    \caption{Average system utilisation on Jetson Orin NX Super 16 GB by \Agiautonomy.}
    \label{tab:utilization_table}
    \vspace*{-12pt}
\end{table}

\subsection{Adaptation of \Agipix}
\begin{figure}[ht!]
        \centering
        \includegraphics[width=0.98\linewidth]{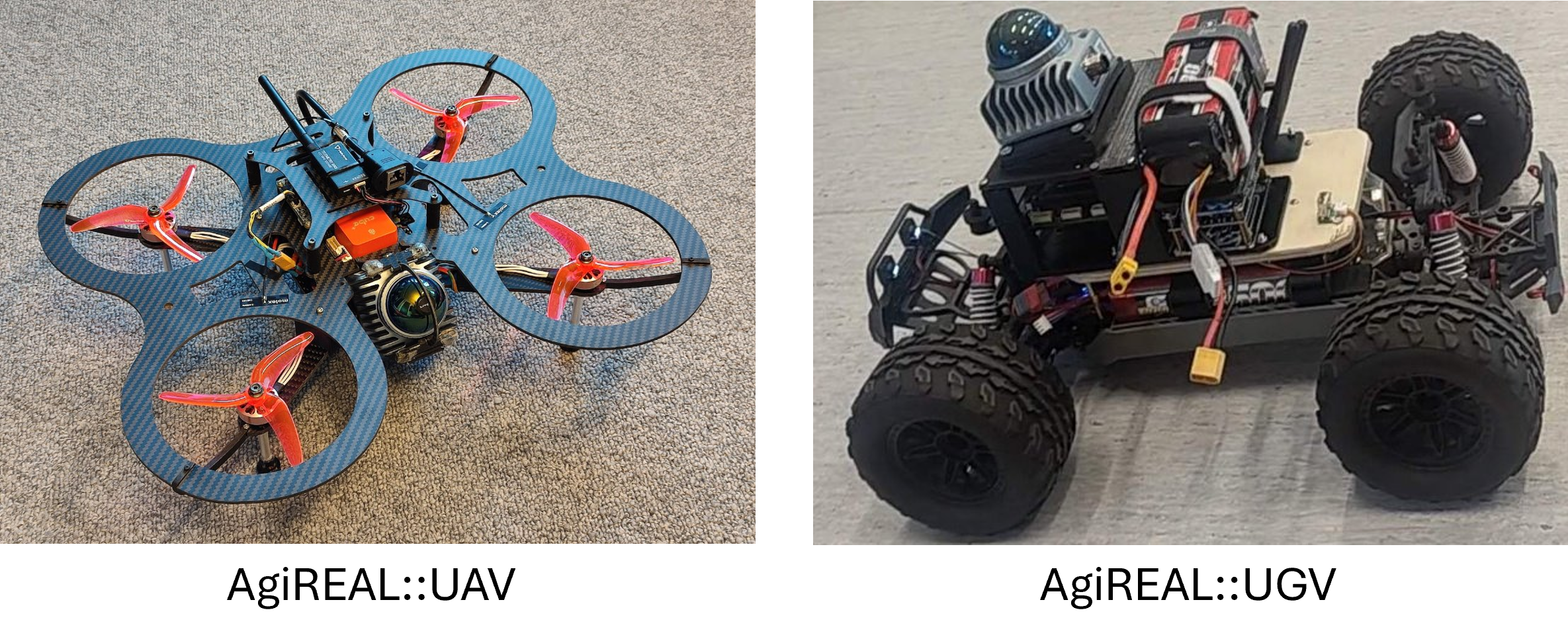}
        \caption{\Agipixs supports heterogeneous hardware deployments. \Agihw::UAV and \Agihw::UGV, shown here, were deployed in ENRICH~2025~\cite{enrich2025raicam}.}
        \label{fig:adaptations}
        \vspace*{-12pt}
\end{figure}
\Agihws hardware design supports interchangeable platform configurations. We provide two variants \Agihw::UAV and \Agihw::UGV as shown in Fig.~\ref{fig:adaptations}, both validated in real-world experiments during ENRICH~2025~\cite{enrich2025raicam}. The architecture shares \Agiautonomys, differing only at the vehicle dynamics and actuation interface, thereby extending the framework beyond a single morphology while maintaining consistent mission and logging workflows.

\section{DISCUSSION}\label{sec:discussion}
\Agipix is designed around a co-design principle: sensing, compute, and autonomy are specified jointly so that the same pipeline can run in both a photorealistic digital twin (\Agisim) and on the real vehicle. The ENRICH~2025 mapping task highlights the practical value of this approach for inspection scenarios, where rapid iteration, reproducible deployment, and robust state estimation are as important as raw flight agility.

\noindent\textbf{Trade-offs.} The compact airframe and protective shell improve close-proximity operation, but the rigid structure increases the risk of frame damage in the event of a hard collision. The compact integration also limits the size and placement of additional sensor modules.\\
\noindent\textbf{Limitations.} The current hardware platform supports connectivity over Wi-Fi~6 and LTE, which can be limiting in extreme environments (e.g., subterranean tunnels or nuclear power plants). Swarm and collaborative mapping capabilities are not yet released.\\
\noindent\textbf{Next steps.} Future work will (i) improve the hardware design by introducing a more compliant impact structure, (ii) expand the digital-twin asset pipeline and sensor-noise models to better match real sites, (iii) improve connectivity by introducing local node-based mesh networking, (iv) introduce cognitive, measurement-induced motion bounds in \Agiuis to improve operator trust and HRI, and (v) add swarm and collaborative mapping capabilities to \Agiautonomy.\\

\section{CONCLUSION}\label{sec:conclusion}
We presented \Agipixs, an open, compact aerial robotics platform for indoor mapping and inspection that bridges simulation and reality through a hardware-synchronized active sensing platform (\Agihw), a containerized ROS~2 autonomy stack (\Agiautonomy), a photorealistic Isaac Sim digital twin (\Agisim), and an operator-facing interface (\Agiui). The platform combines powerful onboard GPU compute with low-level PX4 control to support reproducible deployment and rapid sim-to-real iteration. We report performance focusing on trajectory tracking accuracy, mapping fidelity, and inspection coverage, and we report field validation on the ENRICH~2025 mapping task. By open sourcing all hardware, software, and documentation, we aim to lower financial and engineering barriers and accelerate reproducible research in indoor aerial robotics.

\section*{ACKNOWLEDGMENT}
We acknowledge RAICAM, MSCA HORIZON EU funding. We thank contributors to the open-source ecosystem leveraged by \Agipix.

\bibliographystyle{ieeetr}
\bibliography{ref}

@STRING{jfr     = "J. Field Robot." }

@STRING{ieee    = "Proc. {IEEE}" }

@STRING{icra    = "{IEEE} Int. Conf. Robot. Autom. (ICRA)" }

@STRING{iros    = "IEEE/RSJ Int. Conf. Intell. Robot. Syst. (IROS)" }

@STRING{icuas   = "{IEEE} Int. Conf. Unmanned Aircraft Syst. (ICUAS)" }

@STRING{arxiv   = "ar{X}iv e-prints" }

@inproceedings{enrich2025raicam,
  title={Low-Cost Rapid-Development Air-Ground Robotic Solution for Nuclear Power Plant Inspection},
  author={Tian, Changda and Arachchige, Sasanka Kuruppu and Li, Haichuan and Cardenas, Juan Jose Garcia and Raei, Hamidreza and Dincer, Enes and Kenan, Alperen and Bremner, Paul and Giuliani, Manuel and Neumann, Gerhard and others},
  booktitle={IEEE International Symposium on Safety, Security, and Rescue Robotics},
  year={2025}
}

@article{sanchez2024toward,
  title={Toward fully automated inspection of critical assets supported by autonomous mobile robots, vision sensors, and artificial intelligence},
  author={Sanchez-Cubillo, Javier and Del Ser, Javier and Martin, Jos{\'e} Luis},
  journal={Sensors},
  volume={24},
  number={12},
  pages={3721},
  year={2024},
  publisher={MDPI}
}

@misc{europeanroboticsENRICH2025,
  author       = {{European Robotics}},
  title        = {ENRICH -- The European Robotics Hackathon},
  year         = {2025},
  howpublished = {\url{https://enrich.european-robotics.eu/}},
  note         = {Accessed: 2026-02-03}
}

@misc{marti2023borinot,
  title={Borinot: An Open Thrust-Torque-Controlled Robot for Agile Aerial-Contact Motion Research},
  author={Martí-Saumell, Josep and Duarte, Hugo and Grosch, Patrick and Andrade-Cetto, Juan and Santamaria-Navarro, Angel and Solà, Joan},
  journal={arXiv preprint arXiv:2307.14686},
  year={2023}
}

@misc{djiMatrice400Specs2026,
  author       = {{DJI}},
  title        = {Matrice 400 Technical Specifications},
  year         = {2026},
  howpublished = {\url{https://enterprise.dji.com/matrice-400/specs}},
  note         = {Accessed: 2026-02-03}
}

@misc{skydioX102026,
  author       = {{Skydio, Inc.}},
  title        = {Skydio X10 Technical Specifications},
  year         = {2026},
  howpublished = {\url{https://www.skydio.com/x10/technical-specs}},
  note         = {Accessed: 2026-02-03}
}

@inproceedings{liu2024omninxt,
  title={Omninxt: A fully open-source and compact aerial robot with omnidirectional visual perception},
  author={Liu, Peize and Feng, Chen and Xu, Yang and Ning, Yan and Xu, Hao and Shen, Shaojie},
  booktitle={2024 IEEE/RSJ International Conference on Intelligent Robots and Systems (IROS)},
  pages={10605--10612},
  year={2024},
  organization={IEEE}
}

@article{microswarm2022fast,
author = {Xin Zhou  and Xiangyong Wen  and Zhepei Wang  and Yuman Gao  and Haojia Li  and Qianhao Wang  and Tiankai Yang  and Haojian Lu  and Yanjun Cao  and Chao Xu  and Fei Gao },
title = {Swarm of micro flying robots in the wild},
journal = {Science Robotics},
volume = {7},
number = {66},
pages = {eabm5954},
year = {2022},
doi = {10.1126/scirobotics.abm5954},
URL = {https://www.science.org/doi/abs/10.1126/scirobotics.abm5954},
eprint = {https://www.science.org/doi/pdf/10.1126/scirobotics.abm5954}}

@article{tranzatto2022cerberus,
author = {Marco Tranzatto  and Takahiro Miki  and Mihir Dharmadhikari  and Lukas Bernreiter  and Mihir Kulkarni  and Frank Mascarich  and Olov Andersson  and Shehryar Khattak  and Marco Hutter  and Roland Siegwart  and Kostas Alexis },
title = {CERBERUS in the DARPA Subterranean Challenge},
journal = {Science Robotics},
volume = {7},
number = {66},
pages = {eabp9742},
year = {2022},
doi = {10.1126/scirobotics.abp9742},
URL = {https://www.science.org/doi/abs/10.1126/scirobotics.abp9742},
eprint = {https://www.science.org/doi/pdf/10.1126/scirobotics.abp9742}}

@article{Foehn2022Agilicious,
author = {Philipp Foehn  and Elia Kaufmann  and Angel Romero  and Robert Penicka  and Sihao Sun  and Leonard Bauersfeld  and Thomas Laengle  and Giovanni Cioffi  and Yunlong Song  and Antonio Loquercio  and Davide Scaramuzza },
title = {Agilicious: Open-source and open-hardware agile quadrotor for vision-based flight},
journal = {Science Robotics},
volume = {7},
number = {67},
pages = {eabl6259},
year = {2022},
doi = {10.1126/scirobotics.abl6259},
URL = {https://www.science.org/doi/abs/10.1126/scirobotics.abl6259},
eprint = {https://www.science.org/doi/pdf/10.1126/scirobotics.abl6259}}

@misc{flyabilityElios32022,
  author       = {{Flyability}},
  title        = {Elios 3 – Indoor LiDAR Drone for Industry 4.0},
  year         = {2022},
  howpublished = {\url{https://www.flyability.com/elios-3}},
  note         = {Accessed: 2026-02-03}
}

@article{hayat2016surveyhdi,
  title={Survey on Human-Drone Interaction},
  author={Hayat, Sheraz and Yanmaz, Evsen and Muzaffar, Rashid},
  journal={IEEE Access},
  volume={4},
  pages={8376--8399},
  year={2016},
  publisher={IEEE}
}

@inproceedings{shan2020lio,
  title={Lio-sam: Tightly-coupled lidar inertial odometry via smoothing and mapping},
  author={Shan, Tixiao and Englot, Brendan and Meyers, Drew and Wang, Wei and Ratti, Carlo and Rus, Daniela},
  booktitle={2020 IEEE/RSJ international conference on intelligent robots and systems (IROS)},
  pages={5135--5142},
  year={2020},
  organization={IEEE}
}

@article{xu2022fast,
  title={Fast-lio2: Fast direct lidar-inertial odometry},
  author={Xu, Wei and Cai, Yixi and He, Dongjiao and Lin, Jiarong and Zhang, Fu},
  journal={IEEE Transactions on Robotics},
  volume={38},
  number={4},
  pages={2053--2073},
  year={2022},
  publisher={IEEE}
}

@article{zhao2024adaptive,
  title={Adaptive-lio: Enhancing robustness and precision through environmental adaptation in lidar inertial odometry},
  author={Zhao, Chengwei and Hu, Kun and Xu, Jie and Zhao, Lijun and Han, Baiwen and Wu, Kaidi and Tian, Maoshan and Yuan, Shenghai},
  journal={IEEE Internet of Things Journal},
  year={2024},
  publisher={IEEE}
}

@article{LV-DOT,
  title={LV-DOT: LiDAR-visual dynamic obstacle detection and tracking for autonomous robot navigation},
  author={Xu, Zhefan and Shen, Haoyu and Han, Xinming and Jin, Hanyu and Ye, Kanlong and Shimada, Kenji},
  journal={arXiv preprint arXiv:2502.20607},
  year={2025}
}

@inproceedings{kuruppu2024robust,
  title={Robust Navigation Based on an Interacting Multiple-Model Filtering Framework Using Multiple Tracking Cameras},
  author={Kuruppu Arachchige, Sasanka and Lee, Kyuman},
  booktitle={AIAA SCITECH 2024 Forum},
  pages={1175},
  year={2024}
}

@incollection{furrer2016rotors,
  title     = {RotorS---A Modular Gazebo MAV Simulator Framework},
  author    = {Furrer, Fadri and Burri, Michael and Achtelik, Markus W. and Siegwart, Roland},
  booktitle = {Robot Operating System (ROS): The Complete Reference (Volume 1)},
  editor    = {Koubaa, Anis},
  pages     = {595--625},
  year      = {2016},
  publisher = {Springer International Publishing},
  doi       = {10.1007/978-3-319-26054-9_23}
}

@inproceedings{shah2017airsim,
  title={Airsim: High-fidelity visual and physical simulation for autonomous vehicles},
  author={Shah, Shital and Dey, Debadeepta and Lovett, Chris and Kapoor, Ashish},
  booktitle={Field and service robotics: Results of the 11th international conference},
  pages={621--635},
  year={2017},
  organization={Springer}
}

@inproceedings{song2021flightmare,
  title={Flightmare: A flexible quadrotor simulator},
  author={Song, Yunlong and Naji, Selim and Kaufmann, Elia and Loquercio, Antonio and Scaramuzza, Davide},
  booktitle={Conference on Robot Learning},
  pages={1147--1157},
  year={2021},
  organization={PMLR}
}

@inproceedings{jacinto2024pegasus,
  author    = {Jacinto, Marcelo and Pinto, Jo{\~a}o and Patrikar, Jay and Keller, John and Cunha, Rita and Scherer, Sebastian and Pascoal, Ant{\'o}nio},
  booktitle = {2024 International Conference on Unmanned Aircraft Systems (ICUAS)},
  title     = {Pegasus Simulator: An Isaac Sim Framework for Multiple Aerial Vehicles Simulation},
  year      = {2024},
  pages     = {917--922},
  doi       = {10.1109/ICUAS60882.2024.10556959}
}

@inproceedings{chen2022dlio,
    title        = {Direct LiDAR-Inertial Odometry: Lightweight LIO with Continuous-Time Motion Correction},
    author       = {Chen, Kenny and Nemiroff, Ryan and Lopez, Brett T},
    booktitle    = {2023 IEEE International Conference on Robotics and Automation (ICRA)},
    year         = {2023},
    pages        = {3983--3989},
    doi          = {10.1109/ICRA48891.2023.10160508},
    url          = {https://ieeexplore.ieee.org/document/10160508}
}

@ARTICLE{Baca2021jirs,
  title   = "The {MRS} {UAV} System: Pushing the Frontiers of Reproducible
             Research, Real-world Deployment, and Education with Autonomous
             Unmanned Aerial Vehicles",
  author  = "Baca, Tomas and Petrlik, Matej and Vrba, Matous and Spurny,
             Vojtech and Penicka, Robert and Hert, Daniel and Saska, Martin",
  journal = "J. Intell. Rob. Syst.",
  volume  =  102,
  number  =  1,
  pages   = "26",
  month   =  apr,
  year    =  2021,
  issn    = "1573-0409",
  doi     = "10.1007/s10846-021-01383-5"
}

@ARTICLE{lu2022onmpc,
  author={Lu, Guozheng and Xu, Wei and Zhang, Fu},
  journal={IEEE Transactions on Industrial Electronics}, 
  title={On-Manifold Model Predictive Control for Trajectory Tracking on Robotic Systems}, 
  year={2023},
  volume={70},
  number={9},
  pages={9192-9202},
  doi={10.1109/TIE.2022.3212397}}

@inproceedings{mellinger2011minsnap,
  author    = {Mellinger, Daniel and Kumar, Vijay},
  title     = {Minimum Snap Trajectory Generation and Control for Quadrotors},
  booktitle = {2011 IEEE International Conference on Robotics and Automation (ICRA)},
  year      = {2011},
  pages     = {2520--2525},
  doi       = {10.1109/ICRA.2011.5980409}
}

@misc{song2023perceptionaware,
  title={Learning perception-aware agile flight in cluttered environments},
  author={Song, Yunlong and Shi, Kexin and Penicka, Robert and Scaramuzza, Davide},
  journal={arXiv preprint arXiv:2210.01841},
  year={2022}
}

@article{xu2025navrl,
  author={Xu, Zhefan and Han, Xinming and Shen, Haoyu and Jin, Hanyu and Shimada, Kenji},
  journal={IEEE Robotics and Automation Letters}, 
  title={NavRL: Learning Safe Flight in Dynamic Environments}, 
  year={2025},
  volume={10},
  number={4},
  pages={3668-3675},
  doi={10.1109/LRA.2025.3546069}
}

@article{suomela2025synthetic,
  title={Synthetic vs. Real Training Data for Visual Navigation},
  author={Suomela, Lauri and Arachchige, Sasanka Kuruppu and Torres, German F and Edelman, Harry and K{\"a}m{\"a}r{\"a}inen, Joni-Kristian},
  journal={arXiv preprint arXiv:2509.11791},
  year={2025}
}

@article{suomela2026data,
  title={Data Scaling for Navigation in Unknown Environments},
  author={Suomela, Lauri and Takahata, Naoki and Arachchige, Sasanka Kuruppu and Edelman, Harry and K{\"a}m{\"a}r{\"a}inen, Joni-Kristian},
  journal={arXiv preprint arXiv:2601.09444},
  year={2026}
}

@article{zhou2021egoplanner,
  title={Ego-planner: An esdf-free gradient-based local planner for quadrotors},
  author={Zhou, Xin and Wang, Zhepei and Ye, Hongkai and Xu, Chao and Gao, Fei},
  journal={IEEE Robotics and Automation Letters},
  volume={6},
  number={2},
  pages={478--485},
  year={2020},
  publisher={IEEE}
}

@inproceedings{xuVigoPlanner2026,
  title={Vision-aided UAV navigation and dynamic obstacle avoidance using gradient-based B-spline trajectory optimization},
  author={Xu, Zhefan and Xiu, Yumeng and Zhan, Xiaoyang and Chen, Baihan and Shimada, Kenji},
  booktitle={2023 IEEE International Conference on Robotics and Automation (ICRA)},
  pages={1214--1220},
  year={2023},
  organization={IEEE}
}

@article{Mohta18jfr,
    author      =   {Mohta, Kartik and Watterson, Michael and Mulgaonkar, Yash and Liu, Sikang and Qu, Chao and Makineni, Anurag and Saulnier, Kelsey and Sun, Ke and Zhu, Alex and Delmerico, Jeffrey and Karydis, Konstantinos and Atanasov, Nikolay and Loianno, Giuseppe and Scaramuzza, Davide and Daniilidis, Kostas and Taylor, Camillo Jose and Kumar, Vijay},
    title       =   {Fast, autonomous flight in GPS-denied and cluttered environments},
    journal     =   jfr,
    volume      =   {35},
    number      =   {1},
    pages       =   {101-120},
    doi         =   {10.1002/rob.21774},
    year        =   {2018}
}

@article{holmqvist1994visually,
  title={A visually elicited escape response in the fly that does not use the giant fiber pathway},
  author={Holmqvist, Mats H},
  journal={Visual neuroscience},
  volume={11},
  number={6},
  pages={1149--1161},
  year={1994},
  publisher={Cambridge University Press}
}

@misc{nvidiaIsaacSim2026,
  author       = {{NVIDIA}},
  title        = {Isaac Sim – Robotics Simulation and Synthetic Data Generation},
  year         = {2026},
  howpublished = {\url{https://developer.nvidia.com/isaac/sim}},
  note         = {Accessed: 2026-02-03}
}

@ARTICLE{xu2021CFExplore,
  author={Xu, Zhefan and Deng, Di and Shimada, Kenji},
  journal={IEEE Robotics and Automation Letters}, 
  title={Autonomous UAV Exploration of Dynamic Environments Via Incremental Sampling and Probabilistic Roadmap}, 
  year={2021},
  volume={6},
  number={2},
  pages={2729-2736},
  doi={10.1109/LRA.2021.3062008}}

@INPROCEEDINGS{GiernackiCrazy,
author={Giernacki, Wojciech and Skwierczyński, Mateusz and Witwicki, Wojciech and Wroński, Paweł and Kozierski, Piotr},
booktitle={International Conference on Methods and Models in Automation and Robotics (MMAR)},
title={Crazyflie 2.0 quadrotor as a platform for research and education in robotics and control engineering},
year={2017},
pages={37-42},
doi={10.1109/MMAR.2017.8046794}
}

@INPROCEEDINGS{meier2015px4,
  author={Meier, Lorenz and Honegger, Dominik and Pollefeys, Marc},
  booktitle={2015 IEEE International Conference on Robotics and Automation (ICRA)}, 
  title={PX4: A node-based multithreaded open source robotics framework for deeply embedded platforms}, 
  year={2015},
  pages={6235-6240},
}

@misc{PX4_ROS2_Control_Interface,
  author       = {PX4 Development Team},
  title        = {PX4 ROS 2 Control Interface.},
  year         = {2026},
  howpublished = {\url{https://docs.px4.io/main/en/ros2/px4_ros2_control_interface}},
  note         = {Accessed: 2026-02-09}
}

@misc{PX4_ROS2_bridge,
  author       = {PX4 Development Team},
  title        = {PX4 ROS 2 User Guide.},
  year         = {2026},
  howpublished = {\url{https://docs.px4.io/main/en/ros2/user_guide}},
  note         = {Accessed: 2026-02-09}
}

@misc{kuruppuarachchi2026agi_logger,
  author       = {Kuruppu Arachchige, Sasanka},
  title        = {agi\_logger: Robust ROS 2 data logging for Agipix platform.},
  year         = {2026},
  howpublished = {\url{https://github.com/SasaKuruppuarachchi/agi_logger}},
  note         = {Version v1.0.0. Accessed: 2026-02-09}
}

@misc{foxglove2026,
  author       = {Foxglove Technologies, Inc},
  title        = {Foxglove – The observability stack for Physical AI.},
  year         = {2026},
  howpublished = {\url{https://foxglove.dev/}},
  note         = {Accessed: 2026-02-03}
}

@misc{ika_mqtt_client2026,
  author       = {Institute for Automotive Engineering (ika), RWTH Aachen University},
  title        = {mqtt\_client: ROS 2 MQTT client library.},
  year         = {2026},
  howpublished = {\url{https://github.com/ika-rwth-aachen/mqtt_client}},
  note         = {Accessed: 2026-02-09}
}

@article{DiGregorio2021HumanGCS,
  author  = {Di Gregorio, Marianna and Romano, Marco and Sebillo, Maurizio and Vitiello, Genoveffa and Vozella, Antonio},
  title   = {Improving Human Ground Control Performance in Unmanned Aerial Systems},
  journal = {Future Internet},
  year    = {2021},
  volume  = {13},
  number  = {8},
  pages   = {188},
  doi     = {10.3390/fi13080188},
  url     = {https://doi.org/10.3390/fi13080188}
}

@article{Hu2021CloudStation,
  author  = {Hu, Lian and Pathak, Oindrila and He, Zeyu and Lee, Hyeok and Bedwany, Mohamed and Mica, John and Burke, Peter J.},
  title   = {CloudStation: A Cloud-Based Ground Control Station for Drones},
  journal = {IEEE Journal on Miniaturization for Air and Space Systems},
  year    = {2021},
  volume  = {2},
  number  = {1},
  pages   = {36--42},
  doi     = {10.1109/JMASS.2020.3027520}
}

@article{Zhang2024InterfaceWorkload,
  author  = {Zhang, Wenjuan and Liu, Yunmei and Kaber, David B.},
  title   = {Effect of interface design on cognitive workload in unmanned aerial vehicle control},
  journal = {International Journal of Human-Computer Studies},
  year    = {2024},
  volume  = {189},
  pages   = {103287},
  doi     = {10.1016/j.ijhcs.2024.103287},
  url     = {https://doi.org/10.1016/j.ijhcs.2024.103287}
}

@inproceedings{GarciaCardenas2024CognitiveLoadTeleop,
  author    = {Garc{\'i}a-C{\'a}rdenas, Hei and Tapus},
  title     = {Exploring Cognitive Load Dynamics in Human-Machine Interaction for Teleoperation: A User-Centric Perspective on Remote Operation System Design},
  booktitle = {2024 IEEE/RSJ International Conference on Intelligent Robots and Systems (IROS)},
  year      = {2024},
  pages     = {12204--12211},
  doi       = {10.1109/IROS58592.2024.10802226}
}
\end{document}